\documentclass[lettersize,journal]{IEEEtran}
\usepackage{amsmath,amsfonts}
\usepackage{algorithmic}
\usepackage{algorithm}
\usepackage{array}
\usepackage[caption=false,font=normalsize,labelfont=sf,textfont=sf]{subfig}
\usepackage{textcomp}
\usepackage{stfloats}
\usepackage{url}
\usepackage{verbatim}
\usepackage{graphicx}
\usepackage{cite}
\begin{document}

\title{Estimation of Resistance Training RPE using \\Inertial Sensors and Electromyography}

\author{James Thomas and Johan Wahlström

\IEEEauthorblockA{\vspace{4pt}james.thomas@gmx.co.uk, j.wahlstrom@exeter.ac.uk}}



\maketitle

\begin{abstract}
Accurate estimation of rating of perceived exertion (RPE) can enhance resistance training through personalized feedback and injury prevention. This study investigates the application of machine learning models to estimate RPE during single-arm dumbbell bicep curls, using data from wearable inertial and electromyography (EMG) sensors. A custom dataset of 69 sets and over 1000 repetitions was collected, with statistical features extracted for model training. Among the models evaluated, a random forest classifier achieved the highest performance, with 41.4\% exact accuracy and 85.9\% $\pm1$ RPE accuracy. While the inclusion of EMG data slightly improved model accuracy over inertial sensors alone, its utility may have been limited by factors such as data quality and placement sensitivity. Feature analysis highlighted eccentric repetition time as the strongest RPE predictor. The results demonstrate the feasibility of wearable-sensor-based RPE estimation and identify key challenges for improving model generalizability. 

\end{abstract}


\section{Introduction}

\IEEEPARstart{O}{ver} the last decade, science-supported approaches to strength training optimization have advanced considerably \cite{fisher2011evidence}, which, in turn, has increased the demand for research on resistance training. In parallel, the awareness and application of machine learning (ML) and artificial intelligence in sports have expanded significantly, with demonstrated use cases including exercise classification, rehabilitation monitoring, and performance assessment \cite{Fotos2022}. The convergence of these two areas presents a valuable research opportunity. In this study, we investigate ML methods for using inertial measurements to estimate the rating of perceived exertion (RPE), while also examining the potential role of electromyography (EMG) data during the training phase.

RPE is a key measure in resistance training, quantifying the perceived intensity of exercise. The term ``perceived'' is central, as it implies inherent uncertainty and subjectivity. Several scales exist to measure RPE, such as the Borg Scale (6–20) \cite{borg1982psychophysical}; however, one of the most intuitive and widely used scales is the Borg CR10 scale \cite{borg1982psychophysical}. Defined by Gunnar Borg in 1982, this scale ranges from 1 to 10, where 1 indicates no exertion and 10 represents absolute failure. Generally, values above 6 are recognized as representing difficult exercise \cite{day2004monitoring}. Given its relevance and widespread adoption in resistance training, this paper uses the Borg CR10 scale.

Intensity is widely regarded as a key factor influencing muscular hypertrophy \cite{lasevicius2018effects}, defined as an increase in the cross-sectional area of muscle \cite{schoenfeld2010mechanisms}. In bodybuilding, training intensity is critical for maximizing muscle growth. However, intensity is equally important in strength-focused training, such as powerlifting, as higher intensities promote neural adaptations, improvements in the rate of force development, and strength gains. A detailed understanding of intensity is therefore essential for designing effective training programs that optimize both hypertrophy and strength outcomes. 
If trainees misunderstand the scale, they could be at risk of either poor training results, or significant and lasting injury \cite{vetter2010correlations}. An estimation system mitigates these risks by removing a degree of ambiguity surrounding the measure. This is particularly important in the current digital era, where personal training is increasingly delivered online \cite{passmore2023impact}. In such settings, the ability to remotely monitor and regulate effort has become critical, making an automated system for accurately estimating RPE especially valuable.

A range of studies have investigated the use of ML for estimating RPE. However, most have focused on cardiovascular exercise rather than resistance training. For example, Carey et al. explored exertion estimation in Australian football players using wearable accelerometers, GPS receivers, and heart rate monitors \cite{carey2016predicting}. Whilst numerous studies have investigated exertion estimation using inertial or physiological sensors, research specifically focused on estimating RPE directly from EMG signals, particularly using machine learning and wearable EMG systems, remains limited. This gap is, for example, evident in the PERSIST dataset \cite{albert2022persist}, which integrates inertial sensors, heart rate monitors, and electrocardiography sensors for resistance training, but does not include EMG data. 

In this study, wearable inertial measurement units (IMUs) and surface electromyography (SEMG) sensors are employed to capture movement and muscle activity during resistance training. These sensors were selected due to their non-invasive nature, ease of use, and, in the case of IMUs, their ubiquity in modern wearable devices, making them well-suited for real-world and long-term applications. A novel EMG- and IMU-based dataset of resistance training repetitions is presented, and multiple ML models are evaluated for RPE estimation. EMG data is used only during the training phase to generate labels and inform feature selection. Specifically, extracted EMG features are used to encode labels via dimensionality reduction techniques, and models are trained to estimate these labels using IMU data. During testing, only IMU data is provided as input, reflecting real-world deployment conditions where EMG data is typically not available. Our contributions are threefold: 
\begin{enumerate}
    \item The first investigation of using EMG signals during training for IMU-based RPE estimation is provided, along with benchmarking of multiple ML models for this task;
    \item Key informative features and limitations are identified, offering insights for the design of future wearable-sensor-based exertion monitoring systems;
    \item A novel EMG- and IMU-based resistance training dataset is made publicly available at \\ https://doi.org/10.5281/zenodo.17259403, to support reproducibility and future research.
\end{enumerate}

\section{Data Collection}

\begin{figure}
    \centering
    \includegraphics[width=0.6\linewidth]{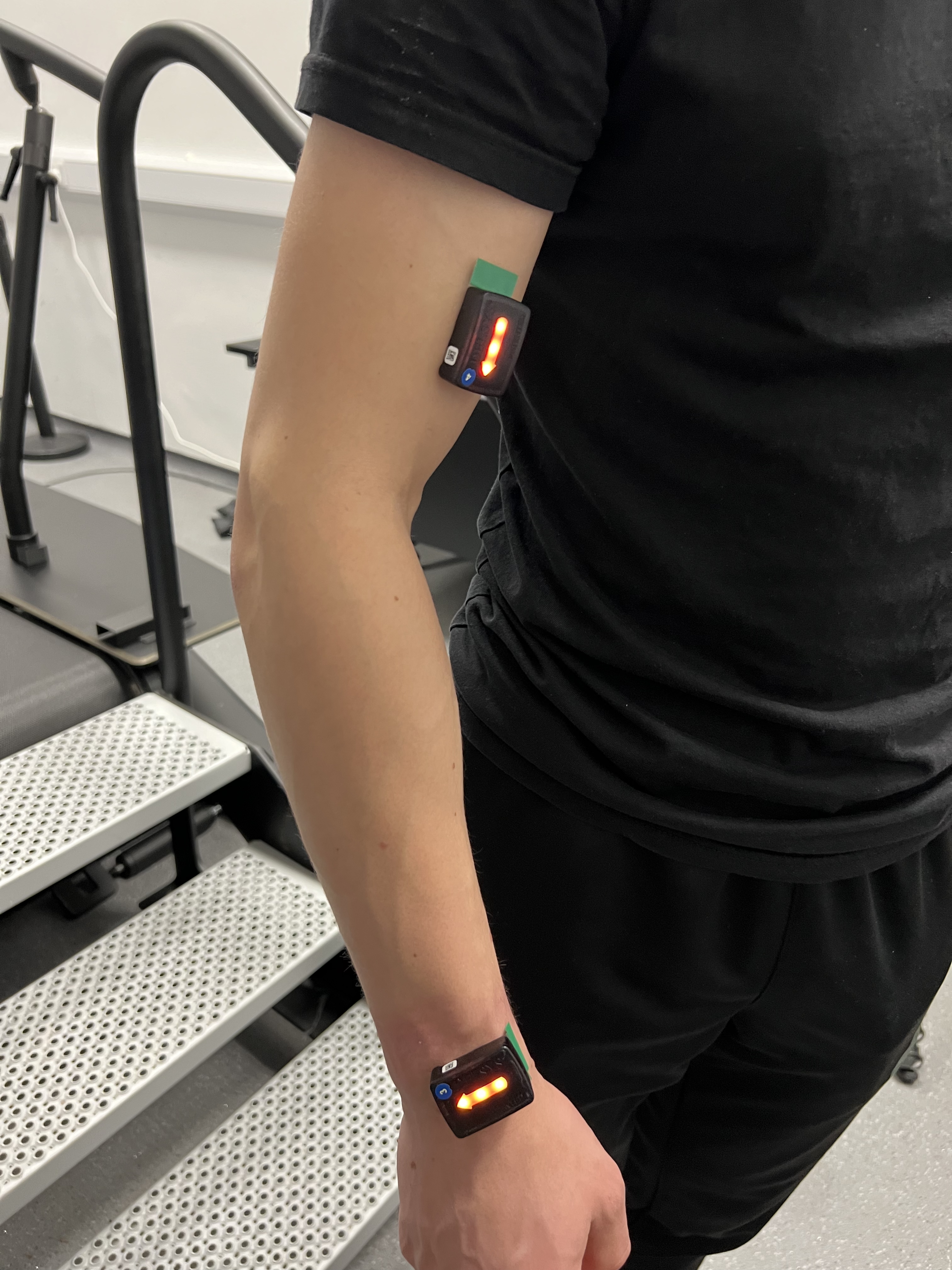}
    \caption{Delsys sensor units attached to a participant.}
    \label{fig:sensors-on-participant}
\end{figure}

Data was collected from five participants, all male. The participants were aged 18–25 and had a minimum of two years of resistance training experience, as well as at least one year of familiarity with the Borg CR10 scale to ensure accurate RPE reporting. Additional inclusion criteria included the absence of recent injuries or physical impairments, guaranteeing consistency in biomechanical movement during data collection. To adhere to anonymity constraints, participants were given pseudo-anonymised IDs created by generating a random letter and then a random 3 digit number. This ID was then tied to data records for identification if participants requested the removal of their data. Further details of participants are presented in Table \ref{tab:participant-info}.

\begin{table}[h!]
\centering
\caption{Participant information with anonymised IDs}
{\small
\begin{tabular}{|c|c|c|c|}
\hline
\textbf{Pseudo ID} & \textbf{Height (cm)} & \textbf{Weight (kg)}  \\
\hline
A321 & 183 & 94 \\
\hline
T417 & 174 & 65 \\
\hline
P714 & 181 & 79 \\
\hline
G998 & 184 & 98 \\
\hline
T456 & 180 & 85 \\
\hline
\end{tabular}
}
\label{tab:participant-info}
\end{table}

\begin{figure}
    \centering
    \includegraphics[width=1\linewidth]{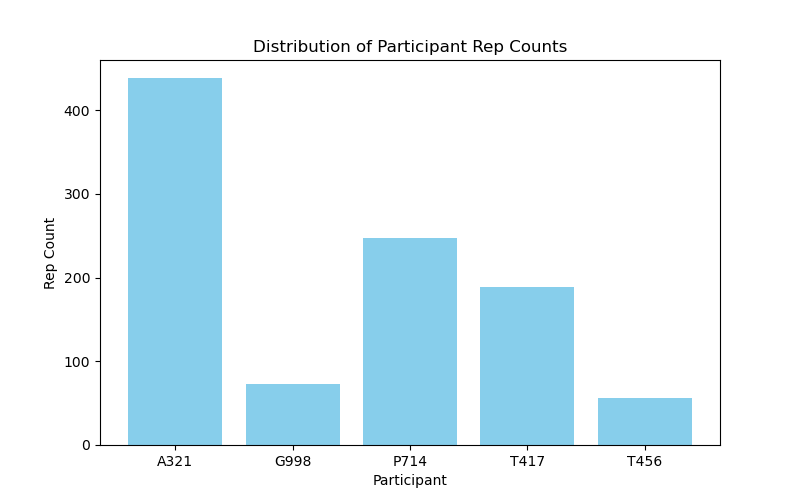}
    \caption{Repetition count per participant.}
    \label{fig:participant-rep-distribution}
\end{figure}

\begin{figure}
    \centering
    \includegraphics[width=1\linewidth]{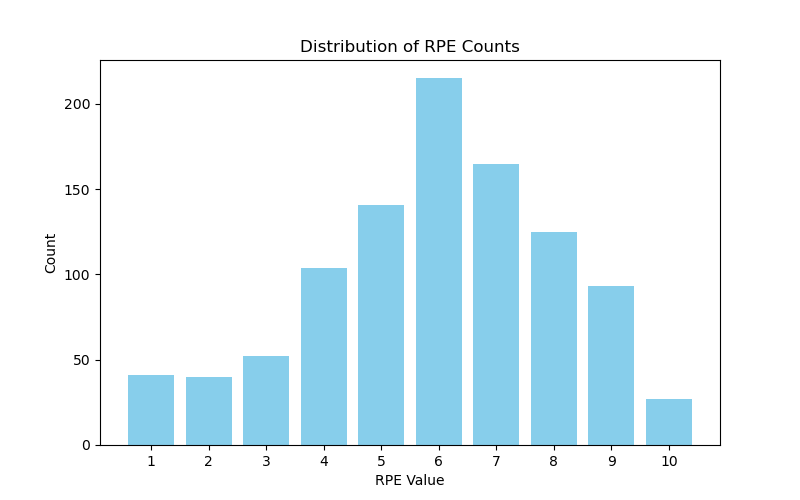}
    \caption{Distribution of collected RPE values.}
    \label{fig:rpe-collection-distribution}
\end{figure}

\begin{figure*}
    \centering
    \includegraphics[width=0.55\linewidth]{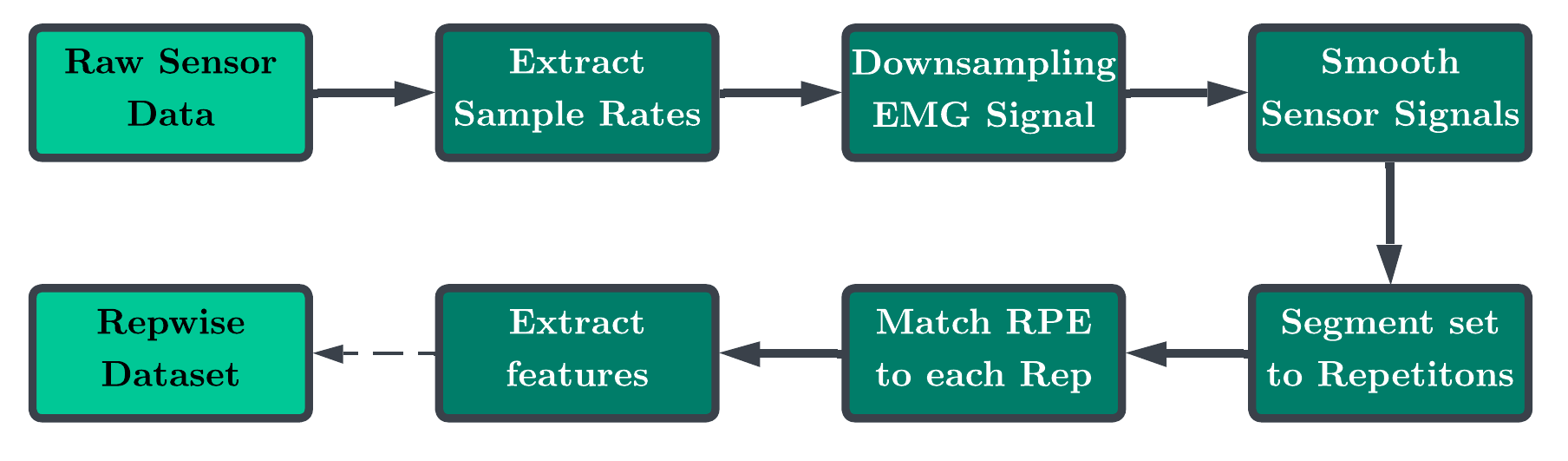}
    \caption{Data processing pipeline.}
    \label{fig:raw-data-pipeline}
\end{figure*}

Data was collected from the single-arm dumbbell bicep curl exercise. This exercise was selected due to its universal nature, meaning it is easy for participants to perform consistently and correctly. The bicep curl also produces a large and easily distinguishable range of motion at the wrist, where wearable exercise trackers are typically worn. EMG and IMU data was recorded using the Delsys Trigno Wireless EMG System \cite{delsys_trigno}, sampling EMG and IMU data at 2148.1 Hz and 370.4 Hz, respectively. Two sensor units were attached to the participant’s bicep and wrist using non-invasive double-sided tape, as shown in Figure \ref{fig:sensors-on-participant}. The wrist unit collected IMU data, as a consumer wearable device would. The bicep unit collected EMG data, since the bicep was the muscle being targeted. The wrist sensor was consistently placed on the outer wrist, positioned horizontally midway between the distal ends of the ulna and radius bones. The bicep sensor was placed on the belly of the biceps muscle, along the midline. During data collection, participants performed repetitions of the exercise and verbally reported their RPE after every repetition. The RPE values were recorded in a spreadsheet and matched to repetitions using unique set IDs, via a Python script. The unique set ID was set using the format of \texttt{userID\_weight\_setnum}, for example, the 9th recorded set with a 15 kg weight for participant A321 was stored as \texttt{A321\_15\_9}.

Participants performed natural sets of bicep curls, completing as many or as few repetitions as they preferred. This approach allowed the data to reflect typical exercise behavior, ensuring a variety of repetition ranges. Dumbbells of 5 kg, 10 kg, and 15 kg were made available, providing a range of difficulties to elicit different levels of exertion. A wide spectrum of RPE values was targeted, which was naturally achieved due to the normally distributed nature of perceived exertion. In total, 69 sets of exercise were collected, with a total of 1003 repetitions. The number of repetitions collected for each participant is shown in Figure \ref{fig:participant-rep-distribution}, and the distribution of RPEs is shown in Figure \ref{fig:rpe-collection-distribution}. 

\section{Methods}

\subsection{Preprocessing}

Figure \ref{fig:raw-data-pipeline} illustrates the process by which raw sensor data is incorporated into the rep-wise dataset. Each set is segmented into individual repetitions, relevant features are extracted, and the resulting data is stored. 

\subsubsection{Sampling Rate Matching}

The sampling rates were dynamically determined for each set to ensure accuracy in case of any misconfiguration. To match the lower IMU sampling rate, the EMG signal was first low-pass filtered using a Butterworth filter and then resampled using polynomial interpolation \cite{de2010filtering}. This ensures that the EMG and IMU signals are aligned, so that each sample corresponds to the same point in time across both sensors, simplifying subsequent processing. Following this, each sensor signal was individually smoothed using a rolling average window \cite{redhyka2015embedded}.

\subsubsection{Repetition Segmentation}

\begin{figure}
    \centering
    \includegraphics[width=0.95\linewidth]{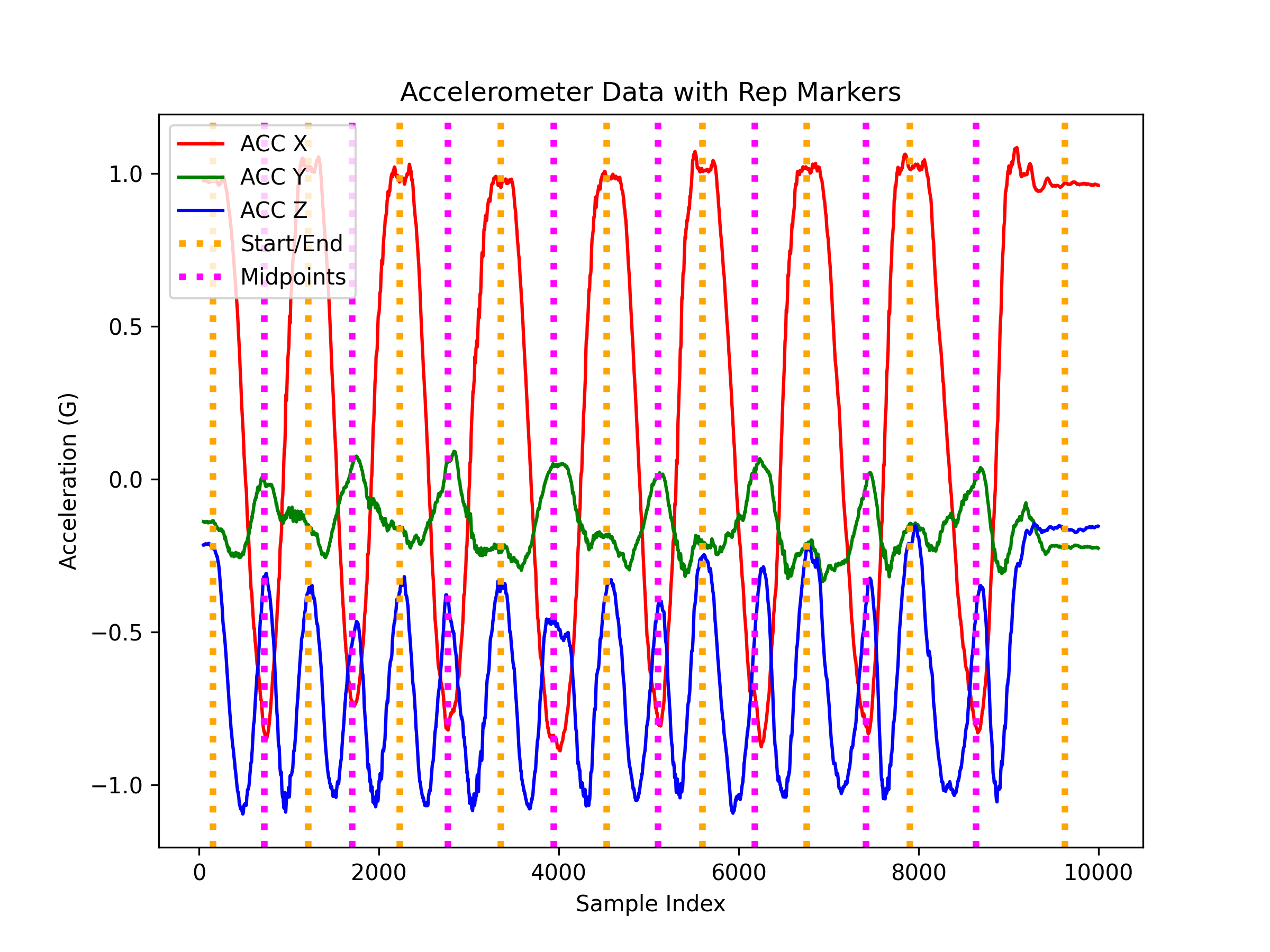}
    \caption{Example of a set marked with repetition boundaries and midpoints.}
    \label{fig:rep-marked-plot}
\end{figure}

Once the sampling rates had been aligned and the signals smoothed, each set of exercises was segmented into individual repetitions. To do this, we first differentiated the accelerometer signal along the axis extending outward from the palm to obtain the jerk, and then identified repetition boundaries and midpoints as instances of zero-crossings in the jerk signal; see Figure \ref{fig:rep-marked-plot}. This procedure can be viewed as a form of peak detection applied to the accelerometer data. A minimum peak distance was imposed to reduce false positives. To ensure accuracy, repetition midpoints were counted and compared against the ground-truth RPE annotations.

It was decided that the end point of one rep would be the starting point of the next, regardless of whether the trainee took a pause in between reps. The break duration and characteristics of movements during these breaks can indicate fatigue, with longer breaks indicating greater fatigue and, in turn, a higher RPE. As such, it was decided that these breaks should be included in repetitions. Reps were then given a unique ID, the ID being the set ID with the rep number from the set. This aids easy rep identification for testing of RNN models. Once each rep had been given an ID, the RPE for each rep was extracted from the RPE file. 

\subsubsection{IMU Feature Extraction}

Following the rep segmentation, features were extracted from the sensor signals for model training. The six IMU data streams enabled the extraction of a wide range of features. Time-based features were computed, including the duration of the concentric (upward) and eccentric (downward) phases of each repetition. For each sensor axis, statistical features such as mean, standard deviation, range, minimum, and maximum were calculated for both concentric and eccentric movements. To quantify the “smoothness” of a repetition, we fitted a polynomial regression to the signal and extracted the coefficient of determination ($R^2$) as a feature. Additionally, jerk was computed, from which statistical features were also derived. Gyroscope features were calculated to capture rotational trends, including mean, standard deviation, and $R^2$. Previous studies have suggested that gyroscope data may be less informative for activity analysis than accelerometer data \cite{alanazi2022effectiveness}, reducing the need for extensive gyroscope feature extraction. In total, 55 IMU features were extracted.

\subsubsection{EMG Feature Extraction} 

EMG features were similarly extracted using statistical metrics, such as mean and root mean square (RMS) to reflect the intensity and power of muscle activation, and variance to capture the variability in muscle power output. The number of zero crossings was computed, as fewer zero crossings can indicate fatigue due to a shift toward lower-frequency activity. Peak amplitude was extracted to represent bursts of muscle activation. In total, nine EMG features were extracted. These features were used solely for generating training labels, which served as targets for ML models using IMU features. In practical applications, IMU data can then be used to estimate these EMG-derived labels, which, in turn, may aid the estimation of RPE.

\subsection{EMG Labeling}

To incorporate EMG training data into a real-world RPE estimation pipeline, models were developed to encode EMG data into labels. Instead of treating the full set of EMG features as ground truth targets—which would require a separate estimation model for each feature—dimensionality reduction was applied to condense the EMG features into a compact label space. These EMG-derived labels capture the essential structure of the muscle activity while limiting the number of models required, as illustrated in Figure 6.

\begin{figure*}
    \centering
    \includegraphics[width=0.55\linewidth]{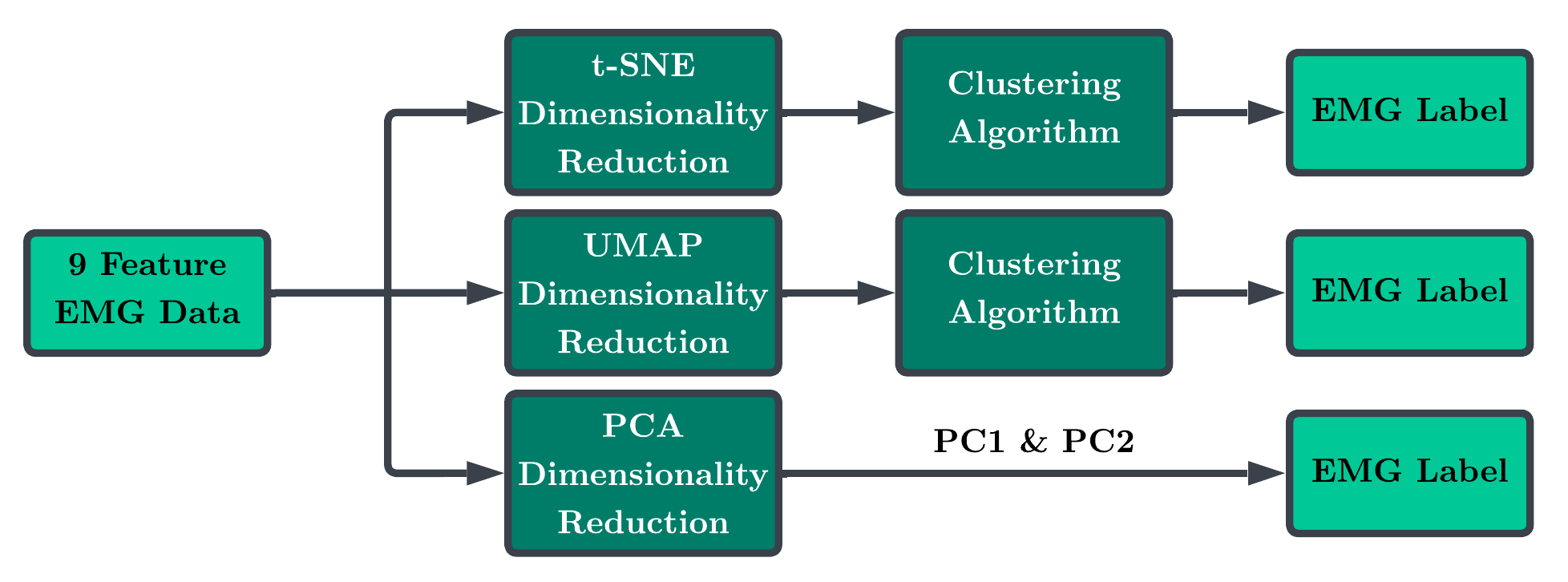}
    \caption{EMG labeling pipeline.}
    \label{fig:emg-label-pipeline}
\end{figure*}

\subsubsection{Dimensionality Reduction}

Three parallel methods for dimensionality reduction were applied: principal component analysis (PCA) \cite{abdi2010principal}, t-distributed stochastic neighbour embedding (t-SNE) \cite{van2008visualizing}, and uniform manifold approximation and projection (UMAP) \cite{mcinnes2018umap}. PCA produces explicit numerical components that capture the variance in the data. The first two components (PC1 and PC2) were selected as labels and can be used directly as continuous targets in regression models. By contrast, t-SNE and UMAP aim to preserve neighborhood structure in the data rather than producing interpretable axes. Their embeddings were therefore used as inputs to a clustering algorithm, with the resulting cluster assignments serving as categorical labels. In this way, PCA yielded continuous features, while t-SNE and UMAP produced categorical features after clustering.

\subsubsection{Clustering Methods}

To transform the t-SNE and UMAP embeddings into discrete labels, clustering was performed using both k-means (KM) and density-based spatial clustering of applications with noise (DBSCAN). DBSCAN did not provide satisfactory results despite hyperparameter tuning, so development proceeded with KM. The number of clusters $k$ was selected as $k=4$ based on optimization of the silhouette score \cite{kodinariya2013review}. To address the class imbalance in the clustered labels, the synthetic minority over-sampling technique (SMOTE) \cite{chawla2002smote} was applied during preprocessing. SMOTE generates synthetic samples of minority classes rather than duplicating existing data, reducing overfitting while preserving class diversity. This step was particularly important given the relatively small dataset, where under-sampling would have further reduced the available training data.

\subsection{EMG Estimation Models}
Once reps had been assigned EMG labels, models were trained to estimate these labels. For PCA, separate regression models were developed to estimate PC1 and PC2 using random forest (RF), support vector regression (SVR), extreme gradient boosting (XGB), and artificial neural network (ANN) architectures. Labels produced by clustering algorithms were discrete categorical labels, and hence, a classification approach was required. To this end, we used RF, logistic regression (LR), support vector machine (SVM), XGB, and ANN architectures. 

\subsection{RPE Estimation Models}
RPE is a discrete categorical label, ranging from 1 to 10. Thus, RPE estimation is naturally a classification problem. However, since the labels are ordinal, regression can also be applied. 
To assess the contribution of EMG features, models will be trained both with and without EMG features, with the latter serving as a baseline. For RPE classification, we will use RF, LR, SVM, XGB and ANN  \cite{davidson2020smartwatch,carey2016predicting,albert2021using}. For RPE regression, we will use RF, support vector regression (SVR), lasso, elasticNet, ridge, ANN, and XGB. 

The approach described above aims to estimate RPE individually at each rep. This approach may overlook longer-term dependencies within a set of exercises. Given the presence of sequence-related dependencies, we will also consider the two main recurrent neural network (RNN) architectures: long short-term memory networks (LSTM) and gated recurrent units (GRU). Since GRUs have fewer gates, they generally train faster than LSTMs, though they may sacrifice some of the representational capacity that LSTMs can offer \cite{chung2014empirical}. If GRUs achieve accuracy comparable to LSTMs, they may therefore be the preferable choice. In real-world applications, RPE estimation would likely need to be performed in real time, making the efficiency advantage of GRUs a particularly valuable feature. These architectures can handle both classification and regression outputs, so both approaches will be evaluated during development. During preprocessing, input samples comprising a fixed number of contiguous repetitions were created. The number of repetitions in each sample was treated as a hyperparameter and optimized during training. The final RPE value within each sample was selected as the representative label for that sample. To mitigate the risk of overfitting caused by overlapping input sequences, a jittering strategy was applied, whereby small amounts of random noise were added to duplicated samples. While this augmentation may slightly reduce accuracy on the training set, it helps improve generalization to unseen data.

\subsection{Data Processing Pipeline}
\begin{figure*}
    \centering
    \includegraphics[width=0.75\linewidth]{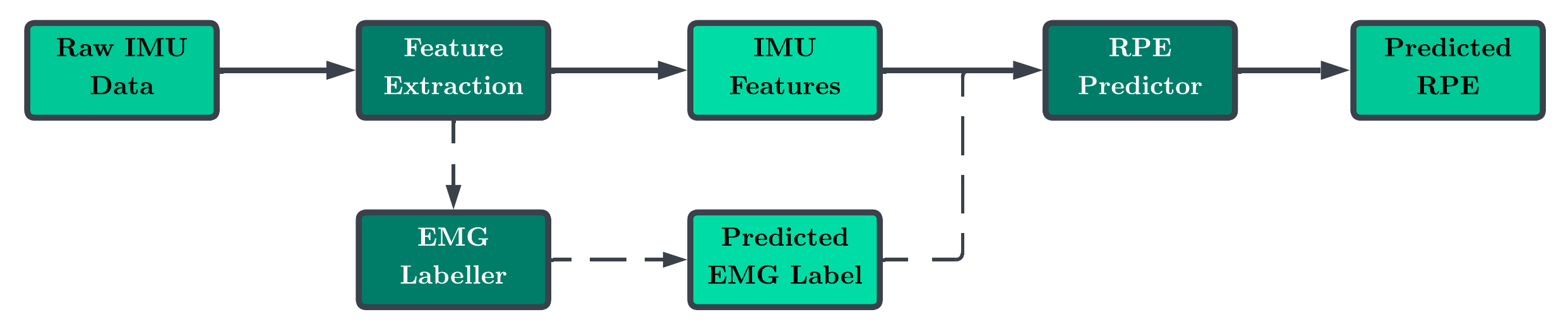}
    \caption{Data processing pipeline for RPE estimation.}
    \label{fig:fresh-data-pipeline}
\end{figure*}

Figure \ref{fig:fresh-data-pipeline} shows the proposed pipeline for estimating the RPE of a new sample. The diagram shows the flow of data, with an optional sub-path for EMG labeling if these features are deemed beneficial. Depending on the final model that is selected, the RPE estimator may accept rep-wise or set-wise input. 

\subsection{Model Evaluation Methodology}\label{sec:model-evaluation-methodology}

A number of evaluation metrics were used. The F1-score was used to evaluate classification model architectures. Accuracy was calculated using a $\pm1$ tolerance, indicating how often the estimated RPE values were within one unit of the ground truth. The $R^2$ score was also used. Root mean square error (RMSE) was the primary metric for evaluating the regression architectures. Throughout, models were evaluated using 4-fold cross-validation. For rep-wise models, folds were created by shuffling individual reps. For RNN models, which operated on overlapping sequences of reps, folds were grouped by set to avoid duplicated reps leaking across train and test splits.

\subsection{Hyperparameter Selection}\label{sec:hyperparam-search-methodology}

Bayesian optimization, that is, iteratively selecting promising parameter combinations to efficiently improve model performance using Bayesian methods \cite{wu2019hyperparameter}, was employed for most hyperparameter tuning. In cases where integrating Bayesian optimization into a model’s workflow proved difficult, random search was used instead.

\section{Experimental Results}
\label{section_experimental_results}
\subsection{EMG Estimation Models}
Tables \ref{tab:emg-pc1} and \ref{tab:emg-pc2} display the performance of the regression models estimating the value of EMG PCA components 1 and 2, respectively. The results for PC1 demonstrate that the XGBoost regressor had the lowest RMSE of 0.6627. The XGB model was therefore selected for producing the estimated labels for RPE model training. The SVR model had a marginally better MAE score, however, the lower RMSE is favoured due to its penalisation of larger errors. RF performed better on the $R^2$ metric, however, the difference of 0.003 is very small. For PC2 estimation, the ANN model performed best on all metrics, with a final RMSE of 0.2230 and a very strong $R^2$ of 0.9446. The superior performance observed on PC2 compared with PC1 is likely due to the lower variation in PC2.

\begin{table}
\caption{Performance on EMG PC1}
    \centering
    \small
    \begin{tabular}{@{}lcccc@{}}
            Model & MAE & MSE & RMSE & $R^2$ \\
            \hline
            \hline
            ANN & 0.3689 & 0.5157 & 0.6922 & 0.5311 \\
            \hline
            RF & 0.3257 & 0.4789 & 0.6648 & \textbf{0.5693} \\
            \hline
            SVR & \textbf{0.3250} & 0.5414 & 0.6991 & 0.5274 \\
            \hline
            XGB & 0.3309 & \textbf{0.4672} & \textbf{0.6627} & 0.5664 \\
    \end{tabular}
    \label{tab:emg-pc1}
\end{table}

\begin{table}
\caption{Performance on EMG PC2}
    \centering
    \small
    \begin{tabular}{@{}lcccc@{}}
         Model & MAE & MSE & RMSE & $R^2$ \\
        \hline
        \hline
        ANN & \textbf{0.1677} & \textbf{0.0499} & \textbf{0.2230} & \textbf{0.9446} \\
        \hline
        RF & 0.1813 & 0.1194 & 0.3314 & 0.8927 \\
        \hline
        SVR & 0.1711 & 0.0539 & 0.2318 & 0.9405 \\
        \hline
        XGB & 0.1830 & 0.1391 & 0.3412 & 0.8843 \\
    \end{tabular}
    \label{tab:emg-pc2}
\end{table}

Tables \ref{tab:emg-tsne-class-results} and \ref{tab:emg-umap-class-results} demonstrate the performance of the EMG classification models estimating the KM generated labels for t-SNE and UMAP representations, respectively. Poorer performing models have been omitted for brevity. As seen in Table \ref{tab:emg-tsne-class-results}, the best model architecture for t-SNE was XGBoost, which performed significantly better than other models across all metrics. Similarly, Table \ref{tab:emg-umap-class-results} shows that for UMAP, RF performed the best, closely followed by XGB. F1-scores of over 0.8, comparable to the accuracy score, show that both models are handling class imbalances well, demonstrating effective use of SMOTE. 

\begin{table}
\caption{Performance on EMG t-SNE Labels}
    \centering
    \small 
    \begin{tabular}{@{}lcccc@{}}
         Model & Accuracy & Precision & Recall & F1 \\
        \hline
        \hline
        RF & 0.8088 & 0.8125 & 0.8088 & 0.8086 \\
        \hline
        XGB & \textbf{0.8147} & \textbf{0.8174} & \textbf{0.8147} & \textbf{0.8147} \\
        \hline
        ANN & 0.8069 & 0.8156 & 0.8069 & 0.8059 \\
    \end{tabular}
    \label{tab:emg-tsne-class-results}
\end{table}

\begin{table}
\caption{Performance on EMG UMAP Labels}
    \centering
    \small
    \begin{tabular}{@{}lcccc@{}}
         Model & Accuracy & Precision & Recall & F1 \\
        \hline
        \hline
        RF & \textbf{0.8396} & \textbf{0.8438} & \textbf{0.8396} & \textbf{0.8400} \\
        \hline
        XGB & 0.8312 & 0.8341 & 0.8312 & 0.8313 \\
        \hline
        ANN & 0.7882 & 0.8082 & 0.7882 & 0.7905 \\
    \end{tabular}
    \label{tab:emg-umap-class-results}
\end{table}

\subsection{RPE Estimation Models}

\subsubsection{Classification Models}
\begin{figure}
    \centering
    \includegraphics[width=0.85\linewidth]{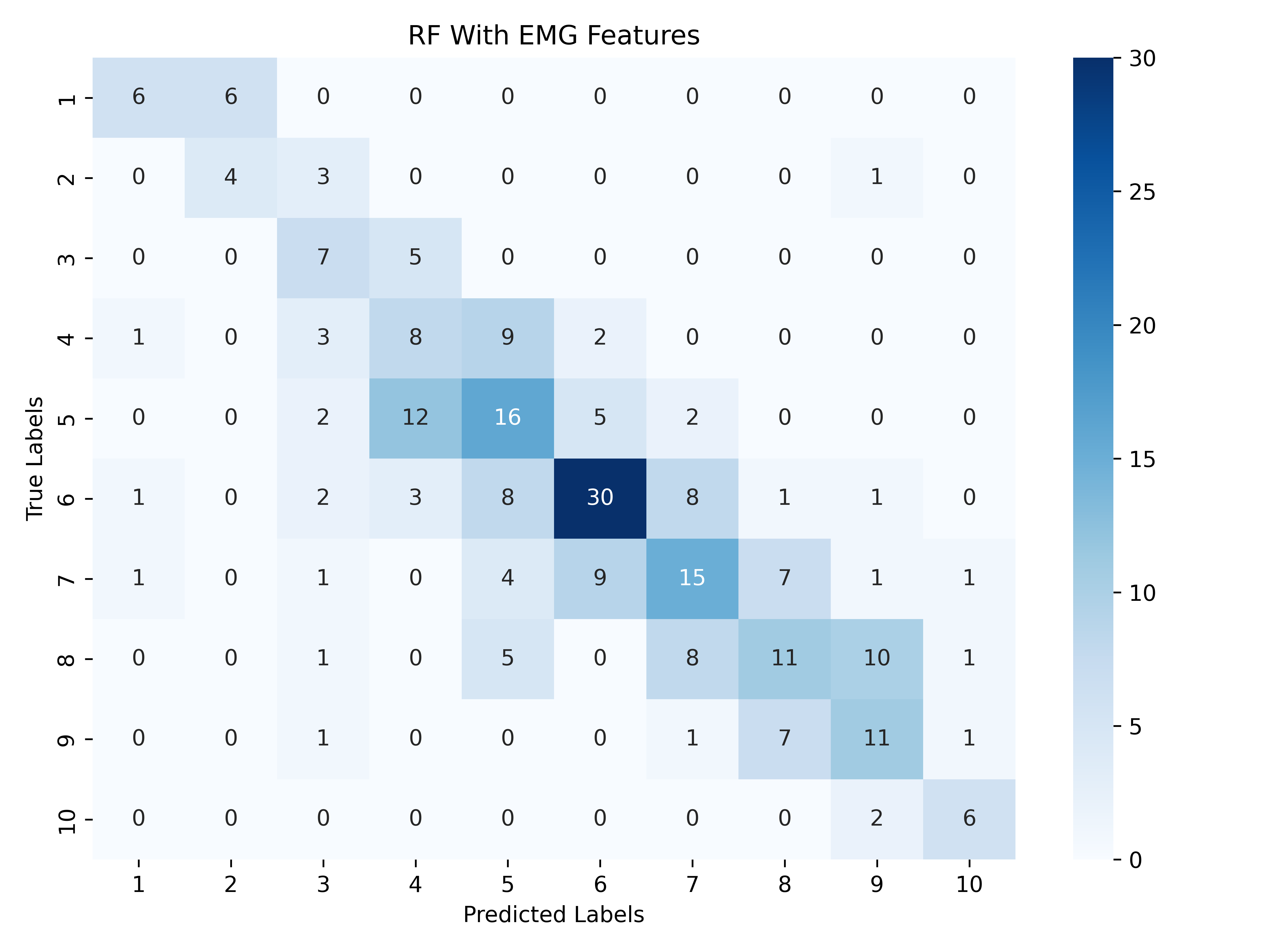}
    \caption{Confusion matrix for the RF classifier with EMG features.}
    \label{fig:rf-cm-emg}
\end{figure}

\begin{figure}
    \centering
    \includegraphics[width=0.85\linewidth]{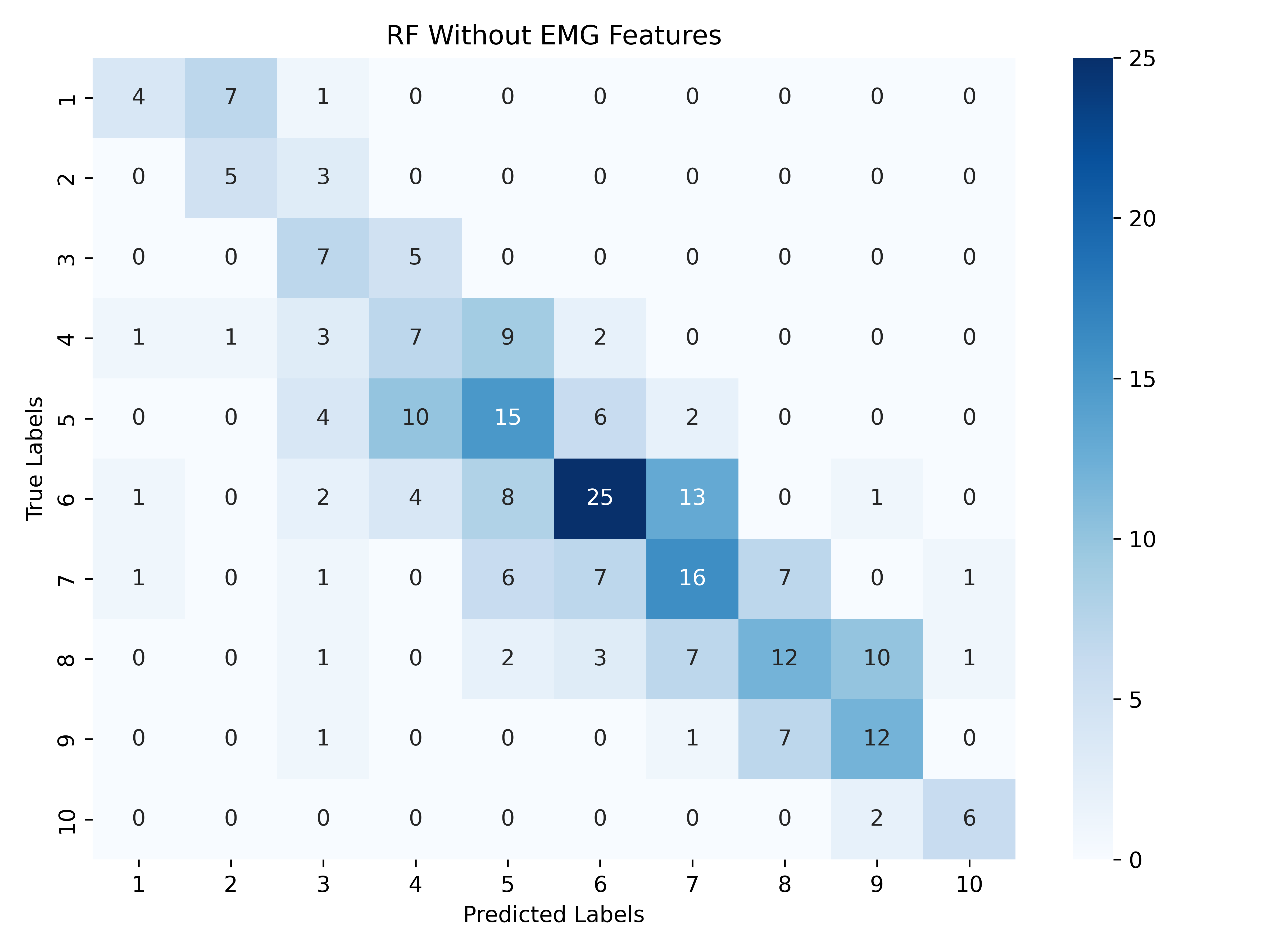}
    \caption{Confusion matrix for the RF classifier without EMG features.}
    \label{fig:rf-cm-no-emg}
\end{figure}

A subset of cross-validation results from the classification model evaluation is presented in Table~\ref{tab:classification-results}. Models with weaker performance have been omitted for brevity. As indicated by the bolded best scores, the RF model with EMG features achieved the highest accuracy. A $\pm1$ accuracy score of 85.9\% (95\% CI: 0.831–0.888) demonstrates that the model is consistently estimating RPE within an acceptable range. Further, the F1-score being marginally higher than the best accuracy demonstrates the model's ability to make a balanced trade-off between precision and recall, as well as reasonable performance on minority class estimates, as supported by the confusion matrices shown in Figures \ref{fig:rf-cm-emg} and \ref{fig:rf-cm-no-emg}. The RMSE of RF without EMG was marginally lower than with, but the majority of metrics favour RF with EMG as the best model. Since it achieved the best performance on five out of six metrics, the RF model with EMG features is considered the most effective classification model for RPE estimation.

\begin{table*}
\caption{Performance of Models for RPE Classification}
\centering
\begin{tabular}{l c c c c c c c}
Model & EMG? & {MAE} & {RMSE} & {Accuracy} & {Acc. ($\pm1$)} & {F1} & {$R^2$} \\
\hline
\hline
ANN & Y & 0.9891 & 1.4896 & 0.3729 & 0.7886 & 0.4074 & 0.3764 \\
\hline
ANN & N & 0.9762 & 1.4510 & 0.3758 & 0.7806 & 0.4059 & 0.3794 \\
\hline
RF & Y & \textbf{0.8116} & 1.2482 & \textbf{0.4138} & \textbf{0.8594} & \textbf{0.4424} & \textbf{0.4249} \\
\hline
RF & N & 0.8225 & \textbf{1.2350} & 0.3968 & 0.8564 & 0.4196 & 0.4085 \\
\hline
XGB & Y & 0.8335 & 1.2361 & 0.4028 & 0.8355 & 0.4270 & 0.4103 \\
\hline
XGB & N & 0.8385 & 1.2676 & 0.4118 & 0.8315 & 0.4234 & 0.4224 \\
\end{tabular}
\label{tab:classification-results}
\end{table*}

\subsubsection{Regression Models}

The results from the regression models are presented in Table~\ref{tab:regression-results}, with weaker-performing architectures omitted for brevity. When calculating classification metrics for regression outputs, the outputs were rounded to the nearest integer to ensure consistent metrics across models for fair comparison. However, the $\pm1$ accuracy metric was computed on unrounded estimates, covering a span of two RPE classes. Unlike the classification setting, where $\pm1$ spans three classes, this results in notably lower $\pm1$ accuracy values. Overall, the XGB regression models achieved the strongest performance, each ranking best on three metrics. The XGB model with EMG features obtained the lowest MAE and RMSE scores. While rounding improved the classification-style metrics slightly, the regression-specific metrics were prioritized, and the XGB model with EMG features is identified as the best-performing regression model. Nonetheless, these regression models perform considerably worse than the classification models on traditional accuracy metrics. Therefore, a classification-based approach to RPE estimation is generally preferred, although regression may be advantageous if capturing subtle differences in exertion is important.

\begin{table*}
\caption{Performance of Models for RPE Regression}
\centering
\small
\begin{tabular}{l c c c c c c c c}
Model & EMG? & {MAE} & {RMSE} & {Accuracy} & {Acc. ($\pm1$)} & {F1} & {$R^2$} \\
\hline
\hline
ANN & Y & 0.8236 & 1.0708 & 0.3769 & 0.6860 & 0.3344 & 0.3908 \\
\hline
ANN & N & 0.8622 & 1.1350 & 0.3758 & 0.6580 & 0.3492 & 0.3985 \\
\hline
SVR & Y & 0.8142 & 1.0654 & 0.3799 & 0.7049 & 0.3611 & 0.4083 \\
\hline
SVR & N & 0.8265 & 1.0682 & 0.3808 & 0.6819 & 0.3375 & 0.4051 \\
\hline
XGB & Y & \textbf{0.7879} & \textbf{1.0439} & 0.4058 & \textbf{0.7059} & 0.3709 & 0.4464 \\
\hline
XGB & N & 0.7927 & 1.0540 & \textbf{0.4177} & 0.7039 & \textbf{0.3807} & \textbf{0.4554} \\
\end{tabular}
\label{tab:regression-results}
\end{table*}

\subsubsection{RNN Models}
The performance of the RNN classification models is presented in Table~\ref{tab:rnn-classification-results}. Interestingly, for both regression and classification tasks, the best-performing models did not include EMG features. Overall, the RNNs underperformed compared to the rep-wise models, with the regression variants performing particularly poorly; these were therefore omitted from the table. This weak performance is likely attributable to the jitter introduced during preprocessing, which may have impaired the integrity of the data. The consistently better performance of classification compared with regression further supports the use of a classification-based approach for RPE estimation. Although the accuracy and F1-scores are substantially lower than those of the rep-wise models, the $\pm1$ accuracy is comparatively less affected, suggesting that even when estimates are incorrect, they tend to fall within the correct region. 

\subsection{Feature Importance for RF RPE Classification}

\begin{figure}
    \centering
    \includegraphics[width=0.85\linewidth]{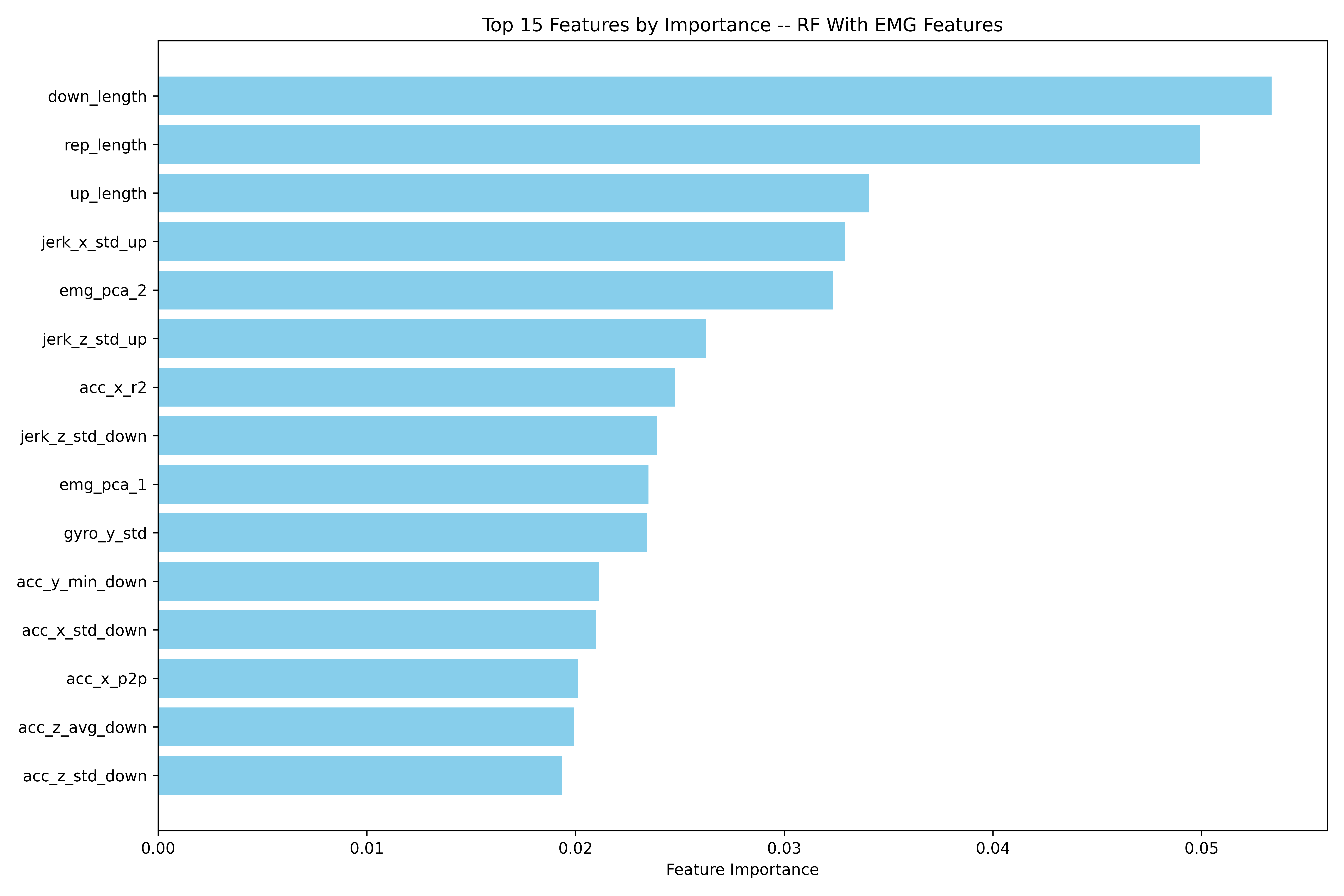}
    \caption{Feature importance for the RF classifier with EMG features.}
    \label{fig:rf-importances-emg}
\end{figure}

Overall, the best-performing model for RPE estimation was the RF classifier using EMG features. Figure~\ref{fig:rf-importances-emg} shows feature importance for this model, computed as the average decrease in impurity each feature provides across all trees. Time-based features were the most influential, followed by jerk features. This indicates that one of the primary determinants of RPE is the time taken to complete a repetition. Interestingly, eccentric duration appears more important than concentric, suggesting that pauses at the end of a rep may serve as a key marker of perceived difficulty. This does not imply that longer repetitions are inherently more difficult; rather, it likely reflects that higher RPEs tend to result in longer repetition times. The association is further supported by a Pearson correlation coefficient (PCC) of $r=0.541$ between rep length and RPE.

\begin{table*}
\caption{Performance of RNN Models for RPE Classification}
\centering
\small 
\begin{tabular}{l c c c c c c c c}
Model & EMG? & {MAE} & {RMSE} & {Accuracy} & {Acc. ($\pm1$)} & {F1} & {$R^2$} \\
\hline
\hline
LSTM & Y & 1.0747 & 1.4853 & 0.3359 & 0.7386 & 0.2662 & 0.3366 \\
\hline
LSTM & N & \textbf{1.0453} & \textbf{1.4453} & 0.3360 & \textbf{0.7444} & 0.2125 & 0.3025 \\
\hline
GRU & Y & 1.0514 & 1.4687 & 0.3327 & 0.7503 & \textbf{0.2903} & \textbf{0.3366} \\
\hline
GRU & N & 1.0914 & 1.5420 & \textbf{0.3396} & 0.7289 & 0.2702 & 0.3351 \\
\end{tabular}
\label{tab:rnn-classification-results}
\end{table*}

\section{Discussion}

The results presented in Section \ref{section_experimental_results} show that classification models, particularly RFs with EMG features, provide the strongest performance for RPE estimation; however, it remains unclear why the inclusion of EMG features yields only modest improvements despite its theoretical relevance to perceived exertion.

\subsection{Impact of EMG features}

Table~\ref{tab:metric-statistics} shows the differences in model performance metrics, computed by subtracting the metrics of each model without EMG features from the corresponding model with EMG features. It should be noted that the architectures were not identical, as different hyperparameters were deemed optimal for different models. In the table, a positive mean indicates that models with EMG tended to perform better, whereas a negative mean indicates that models without EMG performed better. The standard deviations, inherently positive, measure the spread of differences in metrics, i.e., the consistency of the performance differences. Most metrics have positive mean and median values, suggesting that EMG features generally improve performance, albeit marginally. However, the relatively high standard deviations compared to the means highlight substantial variability in these differences, which is further reflected in the minimum and maximum values for each metric. Notably, the maximum values have larger magnitudes than the minimums, indicating that potential gains can exceed potential losses. Accuracy, recall (weighted), F1 (weighted), and $R^2$ all have negative mean values. This likely reflects poor performance on minority classes when EMG features are used, as supported by the contrasting positive averages in the macro metrics. Noise in the EMG data may also contribute to uncertain estimates, particularly for under-represented classes. Overall, the predominance of positive averages supports the use of EMG features as a valid approach to RPE estimation. This conclusion is reinforced by the best EMG-based model achieving a higher F1-score (+0.02) than the best-performing non-EMG model (XGB classifier).

\begin{table*}
\caption{Impact of EMG Features on Model Performance}
\centering
\small
\begin{tabular}{l c c c c c}
Metric & Mean & Median & Std. Dev. & Max & Min \\
\hline
\hline
Accuracy & -0.0040 & -0.0015 & 0.0150 & 0.0170 & -0.0530 \\
\hline
Acc. ($\pm1$) & 0.0072 & 0.0070 & 0.0196 & 0.0296 & -0.0566 \\
\hline
Precision (macro) & 0.0087 & 0.0065 & 0.0315 & 0.0727 & -0.0653 \\
\hline
Recall (macro) & 0.0073 & 0.0035 & 0.0195 & 0.0401 & -0.0380 \\
\hline
F1 (macro) & 0.0057 & 0.0042 & 0.0216 & 0.0537 & -0.0411 \\
\hline
Precision (weighted) & 0.0004 & 0.0009 & 0.0187 & 0.0341 & -0.0565 \\
\hline
Recall (weighted) & -0.0040 & -0.0015 & 0.0150 & 0.0170 & -0.0530 \\
\hline
F1 (weighted) & -0.0025 & -0.0015 & 0.0164 & 0.0205 & -0.0535 \\
\hline
$R^2$ & -0.0115 & 0.0008 & 0.0544 & 0.0504 & -0.1856 \\
\end{tabular}
\label{tab:metric-statistics}
\end{table*}

\begin{figure}
    \centering
    \includegraphics[width=0.85\linewidth]{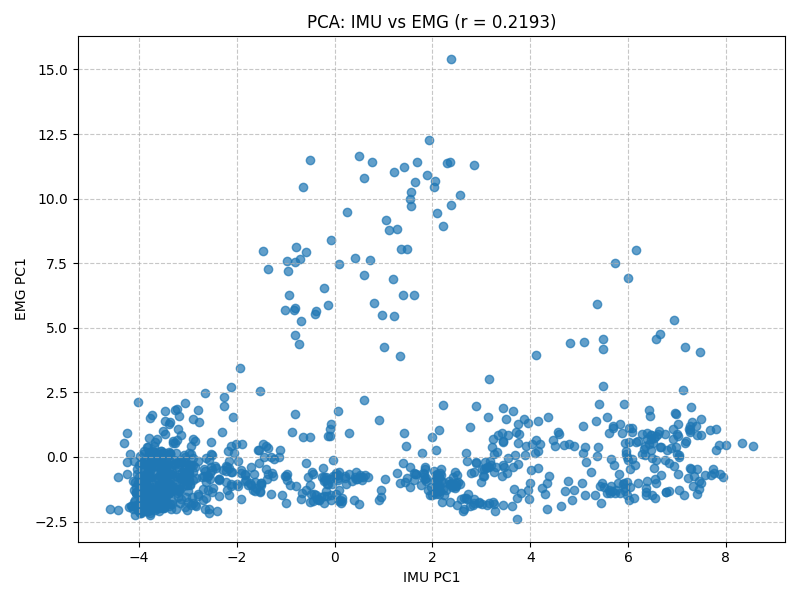}
    \caption{Scatter plot of PC1 values from IMU and EMG feature sets.}
    \label{fig:emg-imu-pca}
\end{figure}

\subsection{Correlations and Physiological Considerations}

The modest performance improvements gained from incorporating EMG training features have several possible interpretations. One possibility is that IMU and EMG data are strongly correlated, such that including EMG provides little additional information beyond what is already captured by the IMU. While strong estimation performance in Tables~\ref{tab:emg-pc1} to~\ref{tab:emg-umap-class-results} might initially support this, further analysis suggests otherwise. Figure~\ref{fig:emg-imu-pca} shows the PC1s from the IMU and EMG feature sets, with a PCC of 0.219, indicating a positive but weak correlation. This implies that the EMG data likely contains additional complexity not captured by the IMU. Another possible explanation for the limited benefit is issues with the underlying EMG data. The selected EMG features may not encompass all relevant information in the raw signals, and irreducible noise could introduce inconsistencies, thereby affecting model performance. This is further supported by the PCC between EMG PC1 and RPE, which is $r=-0.128$, showing a very weak negative correlation. Since EMG measures muscle activation, it might be expected to correlate strongly with perceived exertion. One potential reason this assumption does not hold is the inherent ambiguity of RPE: individuals with different pain tolerances and muscle conditioning may perceive exertion differently. Another factor may be limitations in data quality. Sensors were intended to be consistently placed on the peak of the bicep; however, due to natural variations in muscle anatomy, placement may not have been optimal, potentially resulting in misread signal amplitudes. Additionally, participants with particularly strong forearm or brachialis muscles may have performed a disproportionate amount of work with these muscles rather than the biceps \cite{coratella2023biceps}, further skewing EMG readings. Since prior research indicates a correlation between EMG and RPE \cite{lagally2002perceived}, the weak correlation observed in this study is likely attributable to limitations in data quality and quantity.


\subsection{EMG Feature Extraction}

EMG feature extraction and selection is an area that warrants further exploration. The extracted features were used to encode labels via dimensionality reduction techniques, after which models were developed to estimate these features using inertial measurements. The EMG estimation models achieved reasonable accuracy, demonstrating links between IMU and EMG encodings. The limited benefit of EMG features in the RPE models is unlikely to be due to high EMG estimation errors, as estimation performance was adequate, with low RMSE for PCA labels and classification label accuracy exceeding 80\%. The high feature importance of both PC1 and PC2 in Figure~\ref{fig:rf-importances-emg} further demonstrates the practical effectiveness of these EMG-derived labels. Supervised dimensionality reduction methods, such as linear discriminant analysis (LDA), could potentially enhance the correlation between EMG encodings and RPE. The lower feature importance of the k-means labels, derived from UMAP and t-SNE embeddings, may reflect that these dimensionality reduction methods provided less informative encodings of the EMG data. To investigate this, the best RF architecture was trained using ground truth EMG features instead of estimated labels. This increased model performance to a $\pm1$ accuracy of 86.4\% (+0.5\%) and an absolute accuracy of 41.8\% (+0.4\%). Although these gains are modest, they correspond to improvements of 0.8\% and 2.1\% over the RF model without EMG features, representing a meaningful increase over the baseline. These results suggest that future applications could benefit from estimation models trained directly on EMG features rather than on imputed labels.

\section{Conclusion}
In summary, this project explored the novel use of EMG training data for estimating RPE from inertial sensor input. A variety of models were implemented and evaluated, with Random Forest emerging as the best-performing classifier for RPE using imputed EMG labels. The model achieved a best F1-score of 0.443, indicating reasonable performance given class imbalance, and a $\pm1$ accuracy of 85.9\%, demonstrating strong estimation performance within an acceptable margin. On average, models incorporating EMG features outperformed the baselines, although these gains were marginal.

The modest benefits observed from EMG training data highlight the need for larger and more diverse datasets, ideally on the order of 10,000–20,000 repetitions, with broader participant representation to capture variability in exertion responses. Future work should also investigate alternative strategies for incorporating EMG data, such as focusing on feature selection and extraction of informative raw features rather than reduced encodings. Finally, while rep-wise models outperformed recurrent architectures, the presence of long-term dependencies suggests that RNN-based approaches may prove valuable if trained on larger datasets without reliance on augmentation techniques that compromise temporal integrity.

\bibliographystyle{IEEEtran}

\bibliography{bibliography.bib}

\newpage

\vfill

\end{document}